# Stacking Factorizing Partitioned Expressions in Hybrid Bayesian Network Models

Peng Lin, Martin Neil, and Norman Fenton

*Abstract*— Hybrid Bayesian networks (HBN) contain complex conditional probabilistic distributions (CPD) specified as partitioned expressions over discrete and continuous variables. The size of these CPDs grows exponentially with the number of parent nodes when using discrete inference, resulting in significant inefficiency. Normally, an effective way to reduce the CPD size is to use a binary factorization (BF) algorithm to decompose the statistical or arithmetic functions in the CPD by factorizing the number of connected parent nodes to sets of size two. However, the BF algorithm was not designed to handle partitioned expressions. Hence, we propose a new algorithm called stacking factorization (SF) to decompose the partitioned expressions. The SF algorithm creates intermediate nodes to incrementally reconstruct the densities in the original partitioned expression, allowing no more than two continuous parent nodes to be connected to each child node in the resulting HBN. SF can be either used independently or combined with the BF algorithm. We show that the SF+BF algorithm significantly reduces the CPD size and contributes to lowering the tree-width of a model, thus improving efficiency.

*Index Terms*—hybrid Bayesian networks, partitioned expressions, binary factorization, dynamic discretization, Junction tree

## I. INTRODUCTION

A Bayesian network (BN) is a directed acyclic graphic model representing a joint distribution $p(X)$ with random variables $X = (X_1, \dots, X_m)$ in a factorized way. With conditional independence [1] (CI) assumption of the variables presented in the BN, $p(X)$ can be factorized as the product of the parent-child conditional probability distributions (CPDs), $p(X) = \prod_{i=1}^{m} p(X_i|X_{pa(i)})$ where $pa(i)$ represents the parents of the node $i$. The CPD size equals the number of variables it contains.

A hybrid Bayesian network (HBN) is a BN containing both discrete and continuous variables. Typical CPDs involved in an HBN are partitioned expressions, which are mixture distributions defined on discrete and continuous variables.

This work was supported in part by the National Natural Science Foundation of China (No. 12071308) and the Youth Academic Innovation Team Construction project of Capital University of Economics and Business, China (No. QNTD202109 and No. QNTD202303). We acknowledge Agena Ltd., for software support. *(Corresponding author: Peng Lin).*

Peng Lin is with the School of Statistics, Capital University of Economics and Business, Beijing 100070, China (E-mail: linpeng@cueb.edu.cn).

Martin Neil is with the School of Electronic Engineering and Computer Science, Queen Mary, University of London, E1 4NS, London, UK (E-mail: m.neil@qmul.ac.uk).

Norman Fenton is with the School of Electronic Engineering and Computer Science, Queen Mary, University of London, E1 4NS, London, UK (E-mail: n.fenton@qmul.ac.uk).

General inference methods for HBN involve message passing [1], Monte Carlo [1] and Variational inference [2]. The discrete message passing is usually used for HBN inference because it requires no distributional assumptions for the CPDs and is efficient when the CPD size is not large. To perform discrete message passing, all the continuous variables are discretized [3], [4]. The BN will be converted into a Markov network (MN) using moralization [1], and the CPDs $p(X_i|X_{pa(i)})$ are transformed into factors (or *potential*) $\varphi_a(X_a)$, in which $X_a$ are variables involved in the potential $\varphi_a$. The number of variables in $X_a$ is called the *potential size* $|\varphi_a|$. However, the efficiency of discrete message passing will decrease dramatically with the increased number of continuous variables in a CPD as space complexity increases.

In practice, we can either reduce the potential and/or cluster size to improve the efficiency of discrete message passing. In exact methods, such as the Junction Tree (JT) algorithm [1], [5], a cluster contains a group of nodes and is generated by a triangulation [1] process of the BN. The resulting maximum cluster size measures the space complexity and is fixed. In contrast, in approximate methods, the clusters are flexibly defined by cluster variant algorithms [6], [7]. So the maximum cluster size varies subject to constraints. Efficiency is guaranteed by trading some accuracy. However, in both exact and approximate methods, the maximum cluster size is lower bounded by the maximum potential size. Therefore, reducing the maximum potential size can effectively improve efficiency for discrete message passing algorithms.

Typical methods to reduce the potential size involve factorization from the potential or CPD level. For example, at the potential level, such as Wainwright et al., [8] decomposes the potentials directly of the MN and is used in MN inference tasks. On the other hand, the CPD level-based methods [9], [10], [11] explore independence information in the CPD to reduce the discrete node state combinations. However, these methods cannot apply to continuous nodes, thus not directly applicable to HBNs.

A well-known method of factorizing from the CPD level and is applicable to HBN is Binary Factorization (BF) [12], [13], which factorizes the CPDs into simpler forms to ensure each CPD involves no more than three continuous nodes (two parents and one child). So, after discretization, the maximum potential size can be reduced. BF is generally applicable to CPDs defined by statistical or arithmetic functions over continuous nodes. However, it does not factorize mixture distributions involving both discrete and continuous nodes, such as $f(x) = \sum_{k=1}^{n} \lambda_k f_k(x)$, where $f_k$ is component distribution defined on continuous variables and $\lambda_k$ is mixing weight controlled by discrete variables. This is because BF only factorizes each component distribution $f_k$, where the total number of component distributions is not reduced. Therefore, many continuous parent nodes remain unfactorized.

To resolve this critical issue, we propose a stacking factorization (SF) algorithm by introducing intermediate variables to decompose the densities in the partitioned expressions exactly, ensuring each child node connects to no more than two continuous parents. SF can be either used independently or combined with the BF algorithm. We show SF+BF almost factorize all forms of mixture densities in an HBN. It not only reduces the potential size but also contributes to lowering the maximum cluster size, allowing efficient discrete message passing for HBNs.

## II. CPDs and Their Binary Factorization

### A. Conditional Probability Distributions (CPDs)

Formally, a CPD $p(X_i|X_{pa(i)})$ is:
1) A *node probability table* if all the variables involved are discrete.
2) A *conditional expression* if $X_{pa(i)}$ are all continuous variables, and $p(X_i = x|X_{pa(i)}) \sim f(X_{pa(i)})$, where $f(\cdot)$ is a statistical or arithmetic function.
3) A *conditional partitioned expression* if $X_{pa(i)}$ is composed of a set of discrete variables $\boldsymbol{D}$ and a complementary set of continuous variables $pa(i)\backslash \boldsymbol{D}$. The CPD $p(X_i = x|X_{pa(i)}) \sim f(X_{pa(i)})$ is a mixture distribution such that $f(x) = \sum_{k=1}^n \lambda_k f_k(x)$, in which $\lambda_k$ are the mixture weights controlled by $\boldsymbol{D}$ and $f_k(x)$ is component distribution defined on variables in $pa(i)\backslash \boldsymbol{D}$.

For example, a CPD $p(W = w|X,Y,Z) = Normal(X + Y + Z, 1000)$ is a conditional expression for $W = w$ on all continuous variables. Another CPD:
$p(W = w|X,Y,Z,D) = p(D = d_1) \cdot Arithmetic(X - Y + Z)$
$+ p(D = d_2) \cdot Normal(0.3X + 0.1Y + Z, 1000)$
$+ p(D = d_3) \cdot Student(X \times Y \div Z)$,

is a conditional partitioned expression for $W = w$ on a discrete variable $D$ and continuous variables $X,Y,Z$.

In this paper, we focus on CPDs involving continuous nodes, since after discretization, all CPDs are discrete and the number of continuous nodes involved in each CPD dominates the potential size. We consider reducing the potential size for the following scenarios:
1) A discrete child node with continuous parents.
2) A continuous child node with continuous parents.
3) A continuous child node with both discrete and continuous parents.

So, when these potential sizes are reduced, the maximum potential size in an HBN can be effectively reduced.

### B. Discrete Inference and Binary Factorization

The well-known exact inference is the Junction tree (JT) [5] algorithm. To construct a JT for a BN, we need to conduct moralization [1], node elimination, and triangulation steps. The *moralization* converts a BN into an MN by connecting parent nodes with shared child nodes. Under a node elimination order, the *triangulation* adds edges to the MN to make it chordless. We can obtain a maximal spanning tree as a JT if the triangulation is optimal. Each vertex in the JT corresponds to a cluster composed of a group of variables. The maximum cluster size -1, is called the *tree-width* (t.w.) of a JT. The space complexity of the JT is exponential to tree-width.

We use the Dynamic Discretization Junction Tree (DDJT) [4] algorithm to perform discrete message passing for BNs. DDJT features a two-stage iterative process interchanging between dynamic discretization (DD) and JT. First, the JT is used to query the posterior marginals of the BN. DD runs iteratively with the JT to query the relative entropy errors (REE) between a discretized constant function and the actual likelihood function. The optimal discretization for each continuous variable is found when the REE is reduced to a threshold. DDJT is suitable for discrete inference for hybrid BNs with only discretization errors. Its space complexity is the same as JT.

Next, we explain how the Binary Factorization (BF) [12] algorithm reduces the CPD size. Without loss of generality, we can assume a continuous CPD $p(X_i = x|X_{pa(i)})$ is expressed as an arithmetic expression $f(X_{pa(i)})$ over $X_{pa(i)}$. This expression can be parsed incrementally by smaller expressions that involve only two variables. If the number of parent nodes $|pa(i)| \geq 3$, we can incrementally parse the operands $(+,-,\times,\div,\wedge ...)$ in the expression by introducing intermediate variables. The resulting CPD size will be reduced to involve only three continuous nodes.

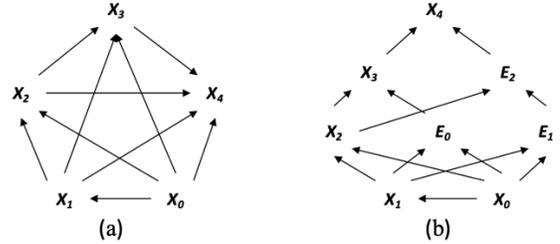

**Fig. 1.** (a) a 5-dimensional complete BN $G$ with all continuous variables; (b) binary factorized BN $G'$ of $G$.

Fig. 1 (a) presents a worst-case BN that is densely connected. Except $X_0$ each node is associated with a conditional expression that is the sum of its parent nodes. The maximum CPD size is 5 in $G$. After BF $G$ is converted to $G'$ in (b), with each intermediate node $E_j, j = 0,1,2$, associated with a conditional expression of the sum of its parent nodes. The expression for node $X_3$ is altered to $p(X_3 = x|X_2, E_0) = Arithmetic(X_2 + E_0)$. The maximum CPD size is 3 in $G'$. Given the arithmetic expressions are commutative, the $G'$ is not necessarily unique. The CPD size is effectively reduced from 5 to 3 in this example.

### C. BF for Partitioned Expressions

The BF has its limitations when applied to conditional partitioned expressions. We use a simple example to illustrate that. In Fig. 2 BN $G$, $D$ is a discrete parent node, while all other nodes are continuous. The continuous parent nodes each have CPDs defined as $Arithmetic(\tau)$ where $\tau$ is a constant.

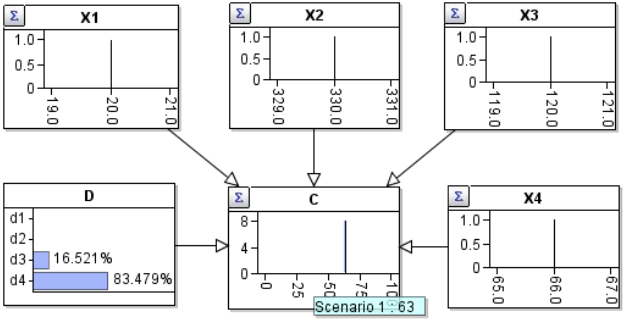

**Fig. 2.** BN $G$ with the child node $C$ observed and associated with a partitioned expression, with marginals.

The CPD for the child node $C$ is defined as:

$$p(C = c|D, X_1, \ldots, X_4) = p(D = d_1) \cdot N(X_1 + X_2 + X_3, 1000)$$
$$+ p(D = d_2) \cdot N(X_2, 1000)$$
$$+ p(D = d_3) \cdot N(X_3, 1000)$$
$$+ p(D = d_4) \cdot N(X_4, 1000)$$

where $N(\cdot)$ is *Normal* distribution.

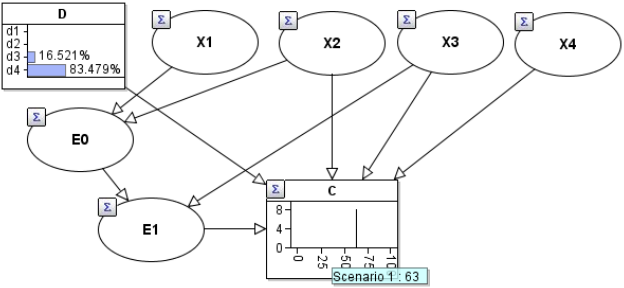

**Fig. 3.** Binary factorized BN $G'$ of $G$ in Fig. 2, with marginals.

Using the standard BN algorithm, as shown in Fig. 3, the CPD for the child node $C$ changes to another partitioned expression:

$$p(C = c|D, X_1, \ldots, X_4) = p(D = d_1) \cdot N(E_1, 1000)$$
$$+ p(D = d_2) \cdot N(X_2, 1000)$$
$$+ p(D = d_3) \cdot N(X_3, 1000)$$
$$+ p(D = d_4) \cdot N(X_4, 1000)$$

where the CPDs for the intermediate nodes $E_0$ and $E_1$ are defined as the sum of their parent nodes given by the BF algorithm. So, the standard BF algorithm only factorized the conditional expression $N(X_1 + X_2 + X_3, 1000)$ leaving the partitioned expression unfactorized. The number of component distributions in the partitioned expression is $K = 4$. If there are $Q$ parent nodes that are binary factorizable and $Q > K$, we gain CPD size reduction for the child node due to BF. Otherwise, the BF cannot reduce the CPD size. In either case, the BF can only reduce the CPD size to $K + 1$ for the partitioned expressions. In this example, $Q = 3$, so the number of parent nodes for $C$ remains unchanged between $G$ and $G'$. The tree-width for both $G$ and $G'$ is 5. In practice, the discrete inference could soon become memory intensive if a cluster contains more than five continuous nodes each discretized into 30 states. So DDJT inference can fail on both $G$ and $G'$ if $X_1, \ldots, X_4$ are distributions rather than a constant $Arithmetic(\tau)$.

### III. STACKING FACTORIZATION

This section develops our stacking factorization (SF) algorithm. The mathematical derivation and algorithm are given in sections III.A and III.B, respectively, and section III.C analysis the space complexity reduction for SF.

*A. Mixture Density Factorization*

Consider the mixture density $f(x) = \sum_{k=1}^{n} \lambda_k f_k(x)$, where $\lambda_k$ are weights of the component distributions $f_k(x)$. Given that $f_k(x)$ are mutually exclusive, and assuming $\lambda_k$ is a constant, we can incrementally construct $f(x)$ by introducing densities $g_1(x), \ldots, g_{n-1}(x)$ in the following way:

$$f(x) = g_{n-1}(x), \tag{1}$$

$$g_k(x) = \alpha_k g_{k-1}(x) + (1 - \alpha_k) f_{k+1}(x), \tag{2}$$

$$\alpha_k = \frac{\sum_{i=1}^{k} \lambda_i}{\sum_{i=1}^{k+1} \lambda_i}, \ (k = 2, \ldots, n-1), \tag{3}$$

where (2) is the recursive relationship between $g_k(x)$ and $g_{k-1}(x)$, with the exit $g_1(x) = \frac{\lambda_1}{\lambda_1 + \lambda_2} f_1(x) + (1 - \frac{\lambda_1}{\lambda_1 + \lambda_2}) f_2(x)$.

**Proof.** of (1) can be arrived at by plugging $k = n - 1$ into (2) and obtaining:

$$g_{n-1}(x) = \alpha_{n-1} g_{n-2}(x) + (1 - \alpha_{n-1}) f_n(x)$$
$$= \frac{\sum_{i=1}^{n-1} \lambda_i}{\sum_{i=1}^{n} \lambda_i} g_{n-2}(x) + \frac{\lambda_n}{\sum_{i=1}^{n} \lambda_i} f_n(x), \tag{4}$$

where (4) can be rewritten as:

$$(\sum_{i=1}^{n} \lambda_i) g_{n-1}(x) = (\sum_{i=1}^{n-1} \lambda_i) g_{n-2}(x) + \lambda_n f_n(x). \tag{5}$$

The relationship of $g_{n-2}(x)$ and $g_{n-3}(x)$ can be derived the same way using (5) to obtain:

$$g_{n-1}(x) = (\sum_{i=1}^{n-2} \lambda_i) g_{n-3}(x) + \lambda_{n-1} f_{n-1}(x) + \lambda_n f_n(x), \tag{6}$$

where the term $g_{n-2}(x)$ is canceled out and $\sum_{i=1}^{n} \lambda_i = 1$.

Therefore, we can iteratively rewrite (6) to obtain $g_{n-1}(x) = \lambda_1 f_1(x) + \cdots + \lambda_n f_n(x) = f(x)$.

To give a simple walkthrough of (1) (2) (3), we assume $n = 4$ and $f(x) = 0.1 \cdot f_1(x) + 0.2 \cdot f_2(x) + 0.3 \cdot f_3(x) + 0.4 \cdot f_4(x)$, thus we introduce $g_1(x), g_2(x), g_3(x)$ with the following settings:

$$g_1(x) = \frac{0.1}{0.1 + 0.2} f_1(x) + \frac{0.2}{0.1 + 0.2} f_2(x),$$

$$g_2(x) = \frac{0.1 + 0.2}{0.1 + 0.2 + 0.3} g_1(x) + \frac{0.3}{0.1 + 0.2 + 0.3} f_3(x),$$

$$g_3(x) = \frac{0.1 + 0.2 + 0.3}{1} g_2(x) + \frac{0.4}{1} f_4(x),$$

where $g_3(x) = f(x)$.

The benefit of using (2) is that we can incrementally reconstruct $f(x)$ using intermediate 'stacked' densities $g_k(x)$, to reconstruct $f(x)$. We call (2) the stacking factorization of the mixture distributions.

The densities in (2) are fixed. To use (2) to factorize the partitioned expressions in a BN, we need to introduce intermediate continuous and discrete control variables. We also need to define the CPDs associated with these random variables, shown in section III.B.

*B. SF for Partitioned Expressions in a BN*

We use an intuitive example to illustrate how the CPDs are specified in the stacking factorization of the partitioned expressions in a BN.

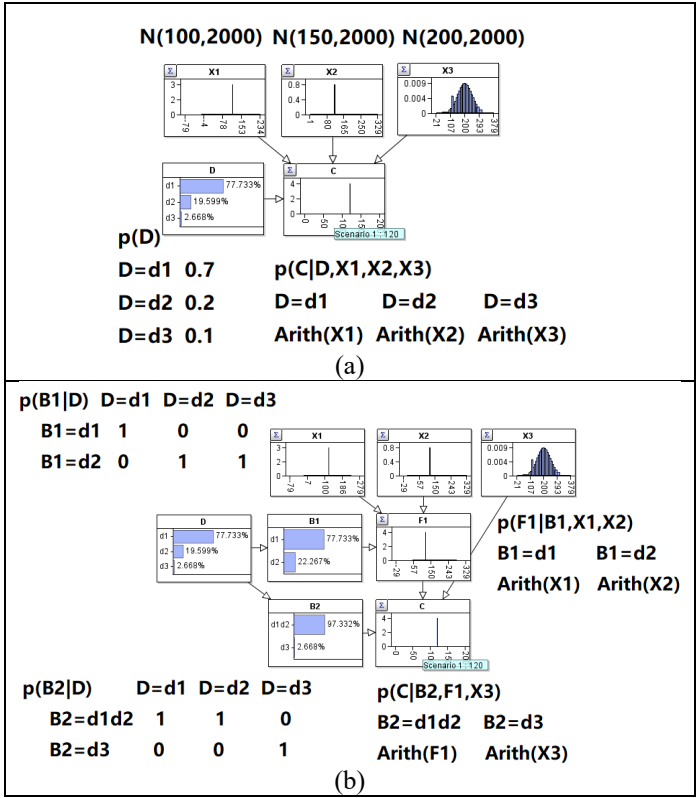

**Fig. 4.** (a) BN $G$ with the CPDs list aside each node, with marginals; (b) BN $G_1$ with the CPDs list aside each intermediate node and the child node $C$, with marginals.

In BN $G$ of Fig. 4. (a), $X_1$ to $X_3$ are *Normal* distributions (short for $N(\cdot)$). The child node $C$ is associated with a partitioned expression with *Arithmetic* (short for $Arith(\cdot)$) component distributions. We set $C = 120$ to validate the posterior marginal probabilities are correct by using the stacking factorization to convert $G$ to $G_1$ (in Fig. 4 (b)).

In $G_1$ the CPDs for the intermediate binary nodes $B_1$ and $B_2$ are set proper ones and zeros to accumulate the relevant probabilities in the parent node $D$, ensuring (3) is held. Taken $B_2$ as an example, its marginal probability at state $B_2 = d_1d_2$ is an accumulation of the marginal probabilities of $p(D = d_1)$ and $p(D = d_2)$, so $p(B_2 = d_1d_2) = 77.733\% + 19.599\% = 97.332\%$, and the complementary probability $p(B_2 = d_3) = 1 - (p(D = d_1) + p(D = d_2)) = 2.668\%$. The CPDs for the intermediate node $F_1$ and the child node $C$ then follow $g_1(x)$ and $g_2(x)$ in (2) directly.

In general, we assume a CPD is represented as a partitioned expression for a child node $C$:
$$p(C = c|D, X_1, \dots, X_n) = p(D = d_1) \cdot f_1(X_1) \\ + p(D = d_2) \cdot f_2(X_2) + \cdots + p(D = d_n) \cdot f_n(X_n), \quad (7)$$

where $D$ is a discrete control node with discrete states $d_1, \dots, d_n$, and $\{X_1, \dots, X_n\}$ is a set of continuous nodes in which each of the nodes is associated with a function $f_j(X_j)$. Equation (7) can be extended to the cases where multiple discrete control nodes are involved, and some continuous nodes are BF applicable.

Following (2), we introduce two sets of intermediate nodes to be added: $\{F_1, \dots, F_{n-2}\}$ and $\{B_1, \dots, B_{n-1}\}$ to factorize the partitioned expression (7), which are continuous and binary discrete nodes respectively. The CPD for $p(B_i|D)$ is defined as:
$$\begin{cases} p(B_i = True|D = d_j) = 1, j \leq i \\ p(B_i = False|D = d_j) = 1, j > i \end{cases}, \quad (8)$$

where $i = 1, \dots, n - 1$ and $j = 1, \dots, n$. Entries in $p(B_i|D)$ other than (8) are set to zeros. Equation (8) ensures that the marginal distribution of $p(B_i)$ is consistent with the weights $\alpha_k$ and $1 - \alpha_k$ in (2), as it must sum over the relevant probabilities in $D$. The CPD for $p(F_i|B_i, F_{i-1}, X_{i+1})$ is then defined as:

$$\begin{cases} p(F_i|B_i = True, F_{i-1}, X_{i+1}) = Arithmetic(F_{i-1}) \\ p(F_i|B_i = False, F_{i-1}, X_{i+1}) = f_{i+1}(X_{i+1}) \end{cases}. \quad (9)$$

So, the intermediate continuous node $F_i$ is associated with a partitioned expression involving only two continuous parent nodes $F_{i-1}$ and $X_{i+1}$. Based on (1), we do not need to add the node $F_{n-1}$ since $F_{n-1} = C$, but simply revise the CPD of the child node $C$ to (9) and reconnect its parent nodes. Therefore, the number of continuous parent nodes for $C$ is reduced from $n$ to 2.

Finally, given (8) can be viewed as an indicator function, we can multiply CPDs (8) to (9) and integrate out $B_i$. We therefore obtain a compact CPD associated to $F_i$:

$$\begin{cases} p(F_i|D = d_j, F_{i-1}, X_{i+1}) = Arithmetic(F_{i-1}), j \leq i \\ p(F_i|D = d_j, F_{i-1}, X_{i+1}) = f_{i+1}(X_{i+1}), j > i \end{cases}. \quad (10)$$

And the CPD for the child node $C$ is changed to:

$$\begin{cases} p(C|D = d_j, F_{n-2}, X_n) = Arithmetic(F_{n-2}), j < n \\ p(C|D = d_j, F_{n-2}, X_n) = f_n(X_n), j = n \end{cases}. \quad (11)$$

Equations (10) and (11) yield our SF algorithm.

**Definition 1.** *Stacking Factorization (SF) is to introduce intermediate continuous nodes $\{F_1, \dots, F_{n-2}\}$ with the associated CPDs defined as (10) to factorize the partitioned expression (7) associated with a child node $C$ in a BN $G$. The resulting BN is $G'$ with $C$ reconnecting to its parent nodes specified by (11). The number of continuous* *parent nodes of C will be reduced from n to 2 (if n > 2).*

**Theorem 1.** *If a BN G is factorized into G' using the SF algorithm, the joint distribution $p(G) = p(G')$.*

**Proof.** Given $p(G)$ equals the product of all the CPDs in $G$. The SF only factorizes the partitioned expressions in $G$, so the CPDs in the form of (7) in $G$ are changed into (10) and (11) in $G'$, with the other CPDs unchanged. The proof is then simplified into proving that (7) can be reconstructed by (10) and (11), using analogy to the proof in section III.A.

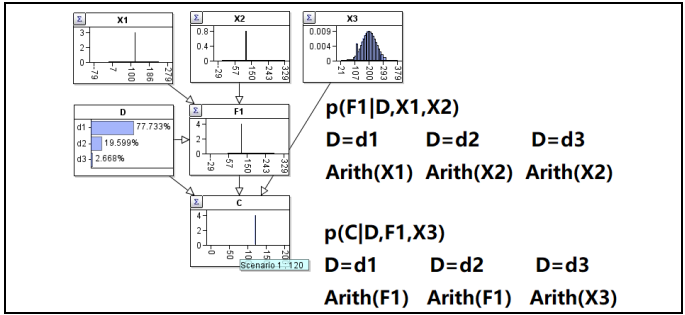

**Fig. 5.** BN $G_2$ with the CPDs list aside for intermediate node $F_1$ and child node $C$, with marginals.

The CPDs in $G_2$ in Fig. 5, follow (10) and (11) to factorize $G$ in Fig. 4 (a), where the relevant conditional expressions are repeated in the partitioned expressions. This is fine because the number of continuous parent nodes will remain two. Both $G_1$ (Fig. 4 (b)) and $G_2$ have reduced the potential size from 5 in $G$ to 4. Theorem 1 is also validated as the marginals are exactly the same in $G$, $G_1$ and $G_2$.

*C. Space Complexity Reduction*

Both BF and SF can reduce the CPDs sizes and can reduce the tree-width (t.w.) of a BN if the largest potential size $|\varphi_a^*|$ dominates the tree-width.

For example, the t.w. for the Fig. 4 BNs is reduced from 4 to 3. This happens when $t.w. + 1 = |\varphi_a^*|$ in the BN. In general, the partitioned expression (7) inherently specifies a converging $V$-structure [1] BN $G$, which is not reducible for the t.w. as the parents are all connected in the moralization process, resulting in $t.w. + 1 = |\varphi_a^*|$. The SF algorithm decomposes the $V$-structure into smaller ones in the factorized BN $G'$. Given that $t.w. + 1 = |\varphi_a^*|$ still holds in $G'$, the t.w. is then reduced from $(n + |D|)$ in $G$ to $(2 + |D|)$ in $G'$, where $|D|$ denotes the number of discrete control nodes.

**Proposition 1**. *The t.w. $\delta$ of a BN G can be reduced to $\delta'$ in its factorized BN G', under the SF algorithm. If the t.w. in G is dominated by the largest potential size of a partitioned expression (7), then $\delta = n + |D| > \delta' \geq 2 + |D|$.*

**Proof.** Proposition 1 defines the upper and lower bound for the t.w. $\delta'$ in $G'$. The proof is straightforward as $3 + |D|$ is the minimum potential size for a partitioned expression, and because the t.w. is lower bounded by the largest potential size $-1$, thus $3 + |D| - 1$ is the lower bound for $\delta'$.

Proposition 1 means that SF guarantees to lower the t.w. from $\delta$ to $\delta'$. However, if the t.w. $\delta$ in $G$ is not dominated by the maximum potential size, such as introduced by the triangulation process, the SF algorithm will not reduce $\delta$ rather than the potential size. Therefore, the efficiency gain is only in the discretization procedure in such cases. Similarly, the BF algorithm also shares proposition 1 without the discrete control nodes $D$. When both SF and BF algorithms are applicable, the lower bound of $\delta'$ is controlled by SF.

## IV. EVALUATIONS

The SF and BF algorithms apply to the three experimental scenarios mentioned in section II.A. We omit the tests for scenarios 1 and 2 as they can be handled solely by BF algorithm [12]. So, we focus on scenario 3 for continuous child nodes having partitioned expressions. The testing environment is Win10, Java 1.8, Agena.AI [13], and i5-9400H. The testing purpose is to evaluate the efficiency gain using the SF+BF algorithm.

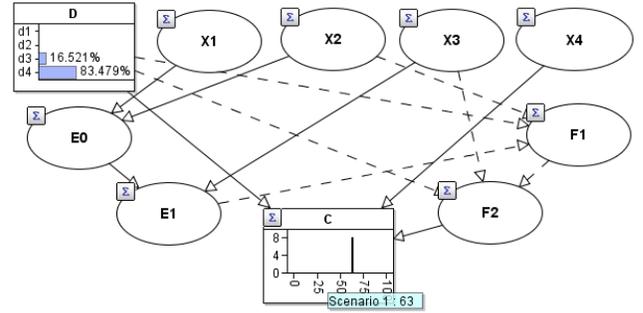

**Fig. 6.** Convert the Fig. 2 BN $G$ into $G'$ using the SF+BF algorithm.

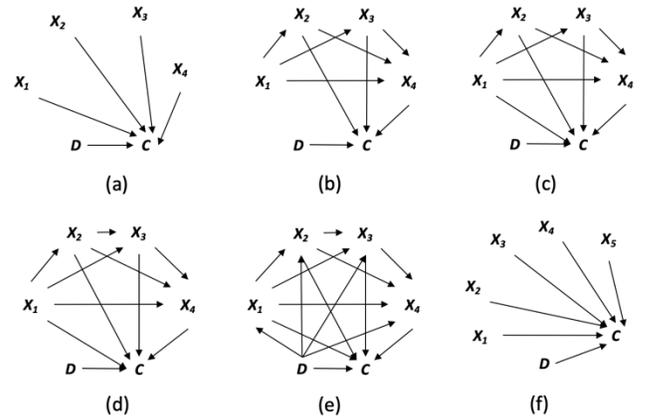

**Fig. 7.** A $2 \times 3$ figure box (a)-(f) for BN $G$ with increased space complexity.

We apply the SF+BF algorithm to the BN $G$ in Fig. 2 and obtain the factorized BN $G'$ in Fig. 6. The dashed lines illustrate the connections for the intermediate nodes $F_1$ and $F_2$ created by the SF algorithm. The other two intermediate nodes $E_0$ and $E_1$ are created by the BF algorithm. All child nodes connect no more than two continuous parent nodes; thus the maximum potential size is reduced from 6 (in $G$) to 4 (in $G'$). The t.w. is also reduced from 5 to 4. If the first component function in the CPD of $X_5$ depends only on $X_1$ (rather than $X_1$, $X_2$, $X_3$) the t.w. will be further reduced to 3. If applying BF without SF for this test, the t.w. will remain unchanged at 5.

Although adding some new nodes to $G'$ introduced extra overheads, the cost is negligible compared to the efficiency gain as the potential size and the space complexity are reduced.

Next, we perform tests covering the worst-case dependent to the independent scenarios, ranging from 6-dimensional to 7-dimensional BNs, as shown in Fig. 7. The space complexity in $G$ increases gradually so we can compare the efficiency gains using the SF+BF algorithm. All tests are run under the DDJT algorithm, which discretizes each continuous node into 20 to 40 discrete states.

In Fig. 7, node $D$ is the discrete control node, and node $C$ is the last continuous child node. Other nodes ($X$) are all continuous. Without loss of generality, we assume the CPDs are partitioned expressions with $Arithmetic$ component functions if both discrete and continuous parent nodes are involved. In this test we also assume CPDs are $sum$ functions if all parent nodes are continuous. Continuous nodes without parents are set as $Normal$ distributions. We index the (a)-(f) BNs $G$ in Fig. 7 from 1 to 6 sequentially and obtain its factorized BN $G'$ (accessible at [14]) by the SF+BF algorithm. The results are given in Fig. 8.

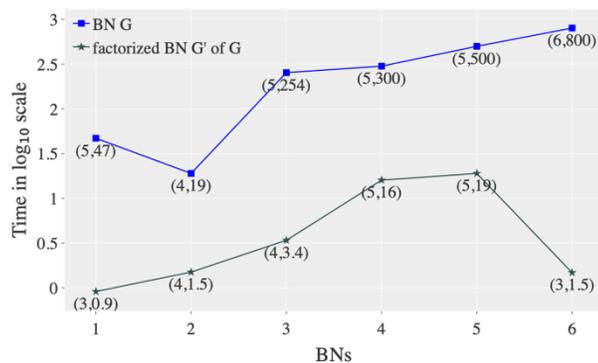

**Fig. 8.** Efficiency comparison for running $G$ and $G'$ using DDJT; the text format aside from the lines is (t.w., seconds).

We set BF on for $G$ in the tests. As all the BNs involve partitioned expressions, the efficiency decreases exponentially with the t.w. in $G$. The BNs $G$ 4, 5, and 6 have yet to be completed as they report a Java heap error after running a few minutes. In contrast, all BNs $G'$ perform robustly, with the most efficiency gains on BNs 1, 3, 6, because both the t.w. and potential sizes are reduced. The t.w. is not reduced for BNs 2, 4, 5 in $G'$ rather than the potential size. However, the efficiency has improved significantly, indicating that reducing the potential size is crucial to further improvement.

## V. CONCLUSION

We presented an SF algorithm that effectively reduces the potential size and t.w. for HBNs involving partitioned expressions. When combined with the BF algorithm, SF can factorize almost all forms of mixture densities, significantly improving efficiency when using discrete message passing for HBNs. Additionally, SF+BF can aid in approximate inference by reducing the lower bound of the cluster size, benefiting other inference methods such as variational inference and belief propagation.

**Peng Lin** received the Ph.D. degree in computer science from Queen Mary, University of London, U.K., in 2015. He is currently an Associate Professor at the School of Statistics, Capital University of Economics and Business, China.

**Martin Neil** is a Professor of Computer Science and Statistics at Queen Mary, University of London, and a Director of Agena, a company that develops Bayesian probabilistic reasoning software and applies it to risky and uncertain problems.

**Norman Fenton** is a Professor of Risk Information Management at Queen Mary, University of London and is also a Director of Agena.